\documentclass[runningheads]{llncs}

\usepackage{comment}

\usepackage{times}
\usepackage{epsfig}
\usepackage{graphicx}
\usepackage{wrapfig}
\usepackage{amsmath}
\usepackage{amssymb}
\usepackage{soul}
\usepackage{color}
\usepackage{pifont}
\usepackage[ruled,vlined]{algorithm2e}
\usepackage{multirow}
\usepackage{booktabs}
\usepackage[skip=5pt]{caption}
\usepackage{subcaption}
\usepackage[mathscr]{euscript}
\usepackage[table,x11names]{xcolor}
\usepackage{enumitem}

\usepackage{hyperref}

\usepackage[accsupp]{axessibility}  

\definecolor{mygray}{gray}{0.95}
\definecolor{myred}{rgb}{1.0, 0.0, 0.0}

\newcommand{\mbf}[1]{\mathbf{#1}}

\newcommand{\y}{\mathbf{y}}

\newcommand{\eg}{\textit{e}.\textit{g}.}
\newcommand\ag[1]{\textcolor{green}{AmirG: #1}} %

\newcommand{\methodname}{SALISA}
\newcommand{\vid}{ImageNet-VID}
\newcommand{\detrac}{UA-DETRAC}

\begin{document}
\pagestyle{headings}
\mainmatter
\def\ECCVSubNumber{6037}  

\title{SALISA: Saliency-based Input Sampling for Efficient Video Object Detection} 

\titlerunning{SALISA}
%
\author{Babak Ehteshami Bejnordi \and
Amirhossein Habibian \and
Fatih Porikli \and
Amir Ghodrati}
\authorrunning{B. Ehteshami Bejnordi et al.}
%
\institute{Qualcomm AI Research\thanks{Qualcomm AI Research is an initiative of Qualcomm Technologies, Inc}\\
\email{\{behtesha,ahabibia,fporikli,ghodrati\}@qti.qualcomm.com}}
\maketitle

\begin{abstract}
High-resolution images are widely adopted for high-performance object detection in videos. However, processing high-resolution inputs comes with high computation costs, and naive down-sampling of the input to reduce the computation costs quickly degrades the detection performance. In this paper, we propose SALISA, a novel non-uniform SALiency-based Input SAmpling technique for video object detection that allows for heavy down-sampling of unimportant background regions while preserving the fine-grained details of a high-resolution image. The resulting image is spatially smaller, leading to reduced computational costs while enabling a performance comparable to a high-resolution input. To achieve this, we propose a differentiable resampling module based on a thin plate spline spatial transformer network (TPS-STN). This module is regularized by a novel loss to provide an explicit supervision signal to learn to ``magnify'' salient regions. We report state-of-the-art results in the low compute regime on the~\vid~and~\detrac~video object detection datasets. We demonstrate that on both datasets, the mAP of an EfficientDet-D1 (EfficientDet-D2) gets on par with EfficientDet-D2 (EfficientDet-D3) at a much lower computational cost. We also show that \methodname~significantly improves the detection of small objects. In particular, \methodname~with an EfficientDet-D1 detector improves the detection of small objects by 77\%, and remarkably also outperforms EfficientDet-D3 baseline.
\keywords{Video object detection, Saliency, Resampling, Efficient Object Detection, Spatial Transformer}
\end{abstract}



\section{Introduction}
The rise in the quality of image capturing devices such as 4K cameras has enabled AI solutions to discover the most detailed video contents and, therefore, allowed them to be widely adopted for high-performance object detection in videos. However, the increased recognition performance resulting from this higher resolution signal comes with increased computational costs. This limits the application of state-of-the-art video object detectors on resource-constrained devices. As such, designing efficient object detection methods for processing high-resolution video streams becomes crucial for a wide range of real-world applications such as autonomous driving, augmented reality, and video surveillance.
To enable efficient video object detection, a large body of works has been focusing on reducing feature computation on visually-similar adjacent video frames~\cite{mao2018catdet,mao2021patchnet,chai2019patchwork} \cite{habibian2021skip,liu2019looking,zhu2017deep,zhu2018towards}. This is achieved by interleaving heavy and light feature extractors~\cite{liu2019looking}, limiting the computation to a local window~\cite{habibian2021skip,chai2019patchwork}, or extrapolating features from a key frame to subsequent frames using a light optical flow predictor~\cite{zhu2017deep,zhu2018towards}. However, these approaches either suffer from feature misalignment resulting from two different feature extractors, or inefficiency in dealing with frequent global scene changes.

\begin{wrapfigure}{r}{0.5\textwidth}
\vspace*{-7.5mm}
\centering
\includegraphics[width=1\linewidth]{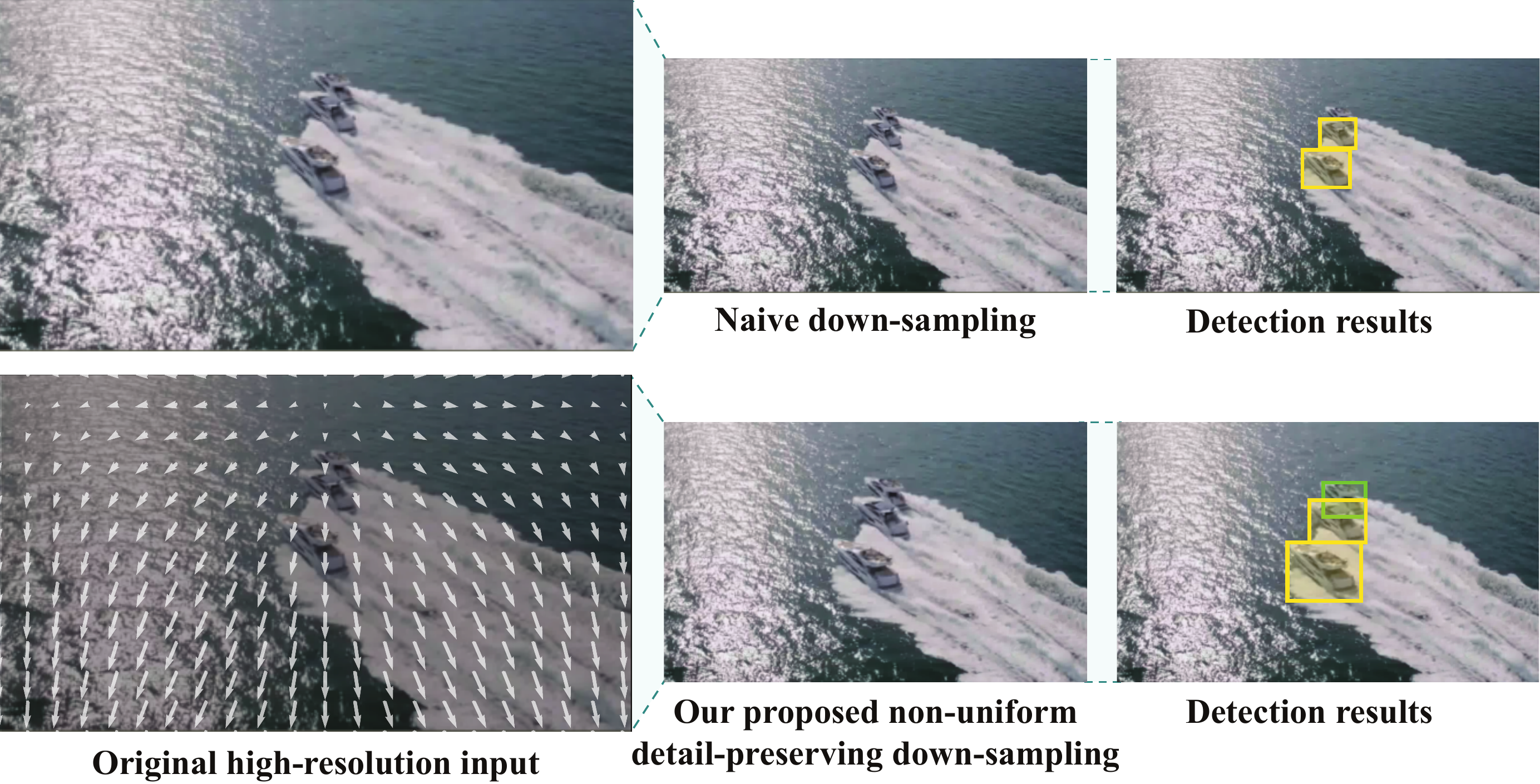}
\caption{An illustration of a non-uniform detail-preserving downsampling of input by \methodname\; leading to improved detection results.}
\label{fig:teaser2}
\vspace*{-5mm}
\end{wrapfigure}


An alternative approach to efficient object detection is to focus on designing lightweight yet highly accurate architectures such as EfficientDet~\cite{tan2020efficientdet}. Recent astounding advances in developing such models have deemed some of the above efficient approaches no longer applicable. For instance, flow-based feature extrapolation might no longer be a proper substitute for existing efficient feature extractors~\cite{tan2020efficientdet,tan2019efficientnet}, as the cost of flow computation is no longer negligible. To be more specific, EfficientDet-D0~\cite{tan2020efficientdet} costs only $2.5$ GFLOPs per frame, while estimating flow by FlowNet-Inception~\cite{dosovitskiy2015flownet} alone costs $1.8$ GFLOPs, translating to $72$\% of the backbone itself~\cite{mao2021patchnet}. However, such efficient architectures may still be expensive when applied to high-resolution video frames. On the other hand, naive down-sampling of the input to reduce the computation costs quickly degrades the performance~\cite{tan2020efficientdet,zoph2020learning}. For example, the performance of EfficientDet-D6 on COCO~\cite{lin2014microsoft} degrades from 52.6\% to 47.6\% when the input is down-sampled by a factor of two~\cite{tan2020efficientdet}.

In this work, we propose \methodname, a novel non-uniform input sampling technique that retains the fine-grained details of a high-resolution image while allowing for heavy down-sampling of unimportant background regions (see Figure~\ref{fig:teaser2}). The resulting detail-preserved image is spatially smaller, leading to reduced computational cost but at the same time enabling a performance comparable to a high-resolution input. Given a sequence of video frames, we first apply a high performing detection model on a high-resolution input at $T=1$ (without resampling). We then generate a saliency map from the detection output to guide the detailed-preserving resampling for the next high-resolution frame. This is achieved via a resampling module that applies a thin plate spline (TPS) \cite{duchon1977splines} transformation to warp the high-resolution input to a down-scaled, detail-preserved one. The resulting resampled frame is then fed to the detector, which consequently has an easier job detecting objects at a lower computational cost.

Our resampling module is based on a thin plate spline spatial transformer network (TPS-STN) \cite{jaderberg2015spatial}. TPS-STN was originally proposed for image classification and used the task loss to train the parameters of STN for digit recognition in MNIST and SVHN. However, adapting this training scheme to object detection in natural images is nontrivial as STN cannot learn to ``magnify'' salient regions without an explicit supervision signal. To address this, we propose a loss term that imposes STN to mimic a content-aware up-sampler. In particular, we use a weighted \textbf{$\ell_2$}-loss between the sampling grid generated by our TPS-STN and the non-parametric attention-based sampler \cite{zheng2019looking} designed for preserving details. Unlike the non-parametric approaches such as attention-based sampler \cite{zheng2019looking} or classical seam carving techniques \cite{setlur2005automatic,avidan2007seam}, our regularized sampling module is fully differentiable, computationally inexpensive, and generates distortion-free outputs.

Our contributions are as follows:
\begin{itemize}
    \item We propose a novel efficient framework for video object detection. Using a saliency map obtained from a previous frame, we perform a non-uniform detail-preserving down-sampling of the current frame, enabling an accurate prediction at a lower computational cost.
    \item To perform the resampling, we develop a fully differentiable resampling module based on a thin plate spline spatial transformer network. We propose a new regularization technique that enables a more effective transformation of the input. 
    \item We report state-of-the-art results in the low compute regime on the~\vid~and~\detrac ~video object detection datasets. In particular, we demonstrate that on both datasets, the mAP of an EfficientDet-D1 (EfficientDet-D2) gets on par with EfficientDet-D2 (EfficientDet-D3) at a much lower computational cost.
    
\end{itemize}




\section{Related Work}

\medskip\textbf{Efficient video Object detection}
A straightforward approach to efficient video object detection is to apply existing efficient object detectors~\cite{ren2015faster,liu2016ssd,bochkovskiy2020yolov4,carion2020end,zoph2020learning,tan2020efficientdet} on a per-frame basis.
However, such an approach does not take the temporal redundancy into account and therefore is computationally sub-optimal for video object detection. In this paper, we specifically use the state-of-the-art cost-effective detection model EfficientDet~\cite{tan2020efficientdet} as our baseline and further extend it for video object detection.

Several methods are proposed to leverage temporal coherency between adjacent frames by tracking previous object detections to skip current detection~\cite{luo2019detect,mao2018catdet}, using template matching to learn patchwise correlation features in adjacent frames~\cite{mao2021patchnet}, limiting the feature computation by processing only a small sub-window of the frames~\cite{chai2019patchwork,habibian2021skip}, using heavy and light networks in an interleaving manner~\cite{liu2019looking}, or efficiently propagating features via a light FlowNet~\cite{zhu2017deep,zhu2018towards}. However, these methods might suffer from tracking errors, misalignments between features, or finding a suitable sub-window. Moreover, with existing efficient backbones~\cite{tan2019efficientnet}, one may find out flow-based techniques no longer yield significant speed-ups, as the cost of flow computation is not negligible. As an alternative, we propose to resample the frame such that it retains the fine-grained details while allowing for heavy downsampling of background areas. The resulting image is spatially smaller, leading to a reduction in computation cost while enabling a performance comparable to a high-resolution input.

\medskip\textbf{Adaptive Spatial Sampling}
One of the major challenges in object detection is to represent and detect fine-grained details in high-resolution images efficiently. One way to tackle this problem is to use hierarchical representations. \cite{xia2016zoom,katharopoulos2019processing,shen2019globally,gao2018dynamic} introduce hierarchical methods to refine the processing of a high-resolution image by adaptively zooming into their proper scales. However, such a hierarchical processing approach makes these methods less suitable for real-time applications.

An alternative approach is to adaptively transform the input such that important fine-grained details are better preserved \cite{jaderberg2015spatial,recasens2018learning,zheng2019looking,gao2020beyond}. The pioneering work of Spatial Transformer Networks (STN) \cite{jaderberg2015spatial} proposes a differentiable module that enables a generic class of input transformations such as affine, projective, and thin plate spline transformations.
While STN works well for MNIST and SVHN datasets, without explicit supervision, it has a hard job of learning effective transformations for complex recognition tasks. Learning-to-zoom \cite{recasens2018learning} uses saliency maps generated by a CNN as guidance to performing a nonuniform sampling that magnifies small details. However, this method causes substantial deformation in the vicinity of the magnified regions, which is particularly harmful when objects overlap or positioned next to each other. Trilinear attention sampling network \cite{zheng2019looking} aims to learn subtle feature representations from hundreds of part proposals for fine-grained image recognition. This technique overcomes the undesirable deformations observed in \cite{recasens2018learning}. However, it is computationally more expensive, non-differentiable, and may still generate undesirable deformations in the background or lower saliency regions. Our method is based on a thin plate spline STN and employs \cite{zheng2019looking} to supervise the STN, allowing it to work on complex datasets while largely eliminating the undesirable distortions caused by \cite{zheng2019looking}.

The adaptive spatial sampling techniques discussed above were primarily designed for image classification tasks. However, optimizing these techniques for downstream tasks such as object detection and semantic segmentation is more challenging. In particular, an undesirable deformation on a non-salient region is unlikely to harm the output prediction of a classification network. At the same time, it can deteriorate the performance of object detection or semantic segmentation model. Jin et al. \cite{jin2021learning} have proposed to use the learning-to-zoom approach \cite{recasens2018learning} for adaptive downsampling of the input for semantic segmentation. To discourage the network from a naive sampling of easy-to-segment regions like background, the authors add an edge loss introduced in \cite{marin2019efficient}. Recently,~\cite{thavamani2021fovea} has proposed a magnification layer based on learning-to-zoom \cite{recasens2018learning} to resample pixels such that background pixels make room for salient pixels of interest. While the major focus of~\cite{recasens2018learning} is on improving object detection accuracy on small objects, we concentrate on increasing efficiency and at the same time improving the performance. 
\vspace{-2mm}
\section{\methodname}
\begin{figure*}[htb!]
\centering
\includegraphics[width=0.95\linewidth]{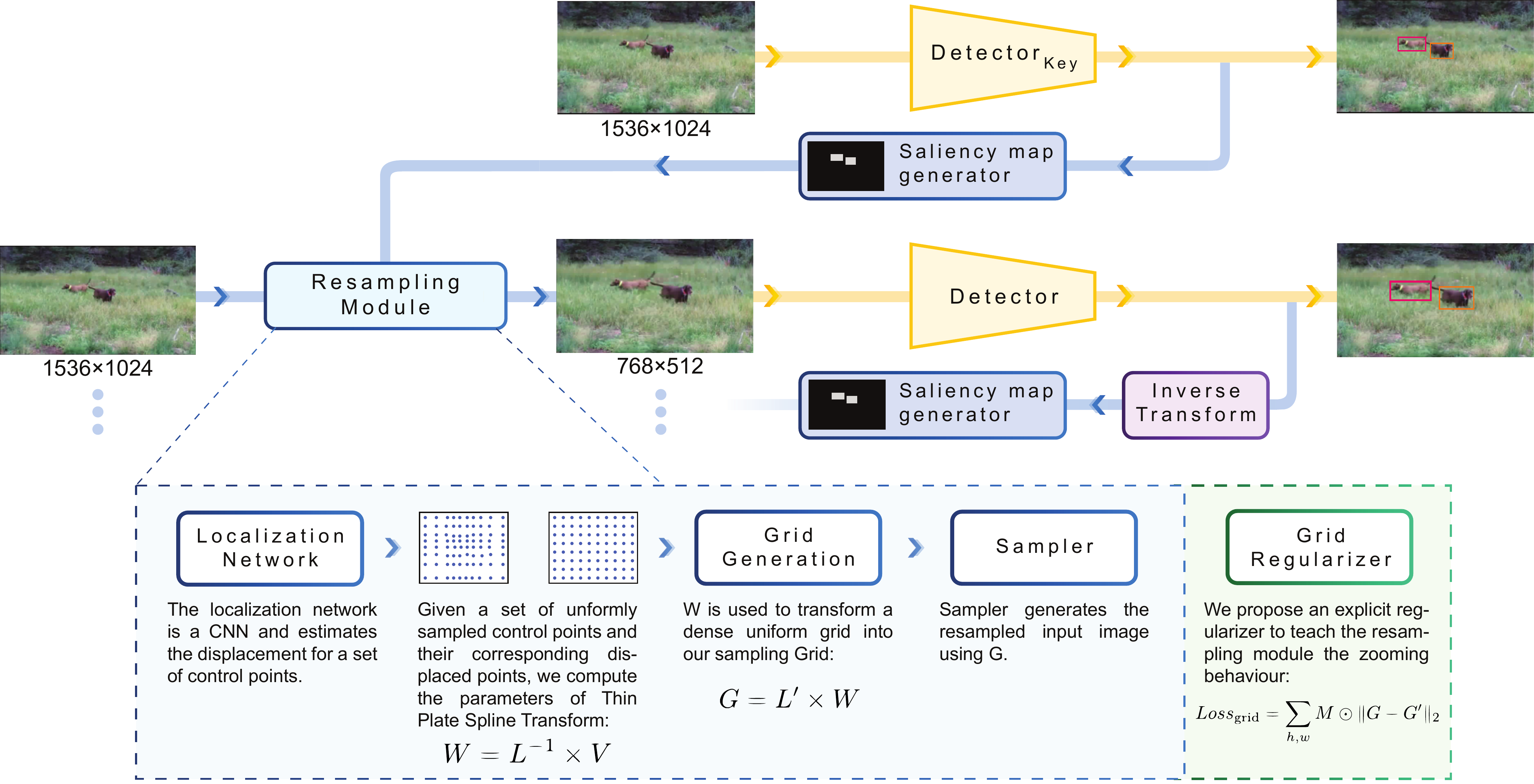}
\caption{{\bf Overview of \methodname.} The first frame, from a set of high-resolution frames, is passed to a high performing detector ($\mathscr{D}_\text{key}$). The saliency map generator uses the prediction output to generate a saliency map. This map and the second high-resolution frame are passed to our resampling module to perform a detail-preserving down-sampling operation. The output of this module is passed to a light detector ($\mathscr{D}$) which is able to perform on par with $\mathscr{D}_\text{key}$ at a much lower computational cost. The output of $\mathscr{D}$ undergoes an inverse transformation to get back to the original image grid before being fed to the saliency map generation for the next frame. This process is continued for processing subsequent frames.}
\label{fig:overview}
\vspace*{-5mm}
\end{figure*}
\vspace{-2mm}
Given a set of high-resolution video frames and their labels $\{\mbf f_i, \y_i\}_{i=1}^N$, we aim to detect the bounding box and category of objects in each frame. Figure~\ref{fig:overview} presents an overview of our proposed SALiency-based Input SAmpling (\methodname) framework for efficient video object detection. \methodname~consists of \textit{i)} two off-the-shelf object detection models $\mathscr{D}_\text{key}$ and $\mathscr{D}$, where FLOPs$_{\mathscr{D}} \ll$ FLOPs${_{\mathscr{D}_\text{key}}}$, \textit{ii) a saliency map generator}, \textit{iii) a resampling module}, and \textit{iv) an inverse transformation module}. At inference, in the first step, we pass the first high-resolution frame ${\mbf f_i}$ (key frame) to a high-performing detection model $\mathscr{D}_\text{key}$. The bounding boxes generated by this model and their corresponding scores are then passed to a saliency map generator to build a global saliency map. This map and the second high-resolution frame ${\mbf f_{i+1}}$ are then passed to our resampling module. The output of the resampling module is a down-sampled detail-preserving image ${\mbf f'_{i+1}}$ which is fed to the light detector $\mathscr{D}$. Due to the nature of this down-sampled image, $\mathscr{D}$ is able to perform on par with $\mathscr{D}_\text{key}$ at a lower computational cost. 
For each of the following frames ${\mbf f_j}$, we generate the saliency map from the detection output of frame ${\mbf f_{j-1}}$ using $\mathscr{D}$. To avoid propagating errors over time, we update the detection output using the strong detector $\mathscr{D}_\text{key}$ at every $S$ frames. In the following sections, we describe the different components of \methodname~in details.
\vspace{-2mm}
\subsection{Saliency map generator}\label{sal_gen}
\vspace{-2mm}
The saliency map generator is a non-parametric detection-to-mask generator, outputting a map corresponding to salient pixels that need to be preserved during resampling. We generate this mask from all the bounding box detections with a score above $\tau$ (See Appendix~\ref{app:tau} for an ablation on the impact of $\tau$ for saliency map generation). The objects with an area $\alpha < 0.5\%$ of the image area are assigned a label of $1$ and the ones with a larger area are assigned a label of $0.5$ (we performed an ablation study on the area parameter $\alpha$ in Section~\ref{sec:exp:ablation}). This will allow our resampling module to focus more on preserving the resolution of smaller objects. The background pixels are labeled as $0$. Note that the saliency values of $0.5$ and $1$ are chosen to make a distinction between large and small objects and the exact choice of saliency values are not critical for performance. We down-sample this saliency map to $128\times128$ before passing it to the resampling module.

\subsection{Resampling module}
Our resampling module is based on a thin plate spline spatial transformer \cite{jaderberg2015spatial}. TPS-STN has three main components: \textit{i)} The localization network, \textit{ii)} The grid generator, and \textit{iii)} the sampler. 

\noindent\textbf{Localization network.} Our localization network is a VGG-style \cite{simonyan2014very} architecture consisting of 10 convolutional and 2 fully connected layers ($0.06$ GFLOPs and $739$k parameters). This network gets the saliency map as input and estimates the displacement of a set of $N=256$ control points defined on a $16\times16$ grid in a Euclidean plane.

\noindent\textbf{Grid generator.} The grid generator is responsible for producing the sampling grid and works as follows. Given a set of $N$ control points sampled uniformly on a 2D grid $\overset{.}{P}\in \mathbb{R}^{N\times 2}$ and their corresponding displaced control points
$\overset{.}{V}\in \mathbb{R}^{N\times 2}$
provided by the localization network, we solve a linear system to derive the parameter $W \in \mathbb{R}^{(N+3)\times 2}$ of TPS as follows:
\begin{equation}
W
=
\underbrace{\begin{bmatrix}
    K & P  \\
    P^T & O
\end{bmatrix}}^{~~~~~~~~~~~~~~~~~~~-1}_\text{$L$}
\times
V, \quad\quad
P=[\textbf{1}, \overset{.}{P}], \quad\quad
V=
\begin{bmatrix}
    \overset{.}{V}  \\
    \textbf{0}
\end{bmatrix}
\end{equation}

where the submatrix $K \in \mathbb{R}^{N\times N} $ is defined as $K_{ij} = U(\|\mbf{p}_i, \mbf{p}_j\|)$ where $\mbf{p}\in \overset{.}{P}$ and $U(r)=r^2log(r)$ is the radial basis kernel. $O \in \mathbb{R}^{3\times 3}$, and $\textbf{0} \in \mathbb{R}^{3\times 2}$ are submatrices of zeros and $\textbf{1} \in \mathbb{R}^{N\times 1}$ is submatrix of ones. Note that one can precompute $L \in \mathbb{R}^{(N+3) \times (N+3)}$ and its inverse. We refer the reader to the Appendix~\ref{app:TPS} for the detailed overview of the algebraic crux of the thin plate method. 

Once we estimate $W$, we can conveniently apply the deformation to a dense uniform grid to obtain the sampling grid $G$, as follows: 
\begin{equation}
G = L' \times W ,
\end{equation}
where $L'$ is computed similarly to $L$ but with dense points. 

\begin{wrapfigure}{rt}{0.48\textwidth}
\vspace*{-8mm}
\centering
\includegraphics[width=1\linewidth]{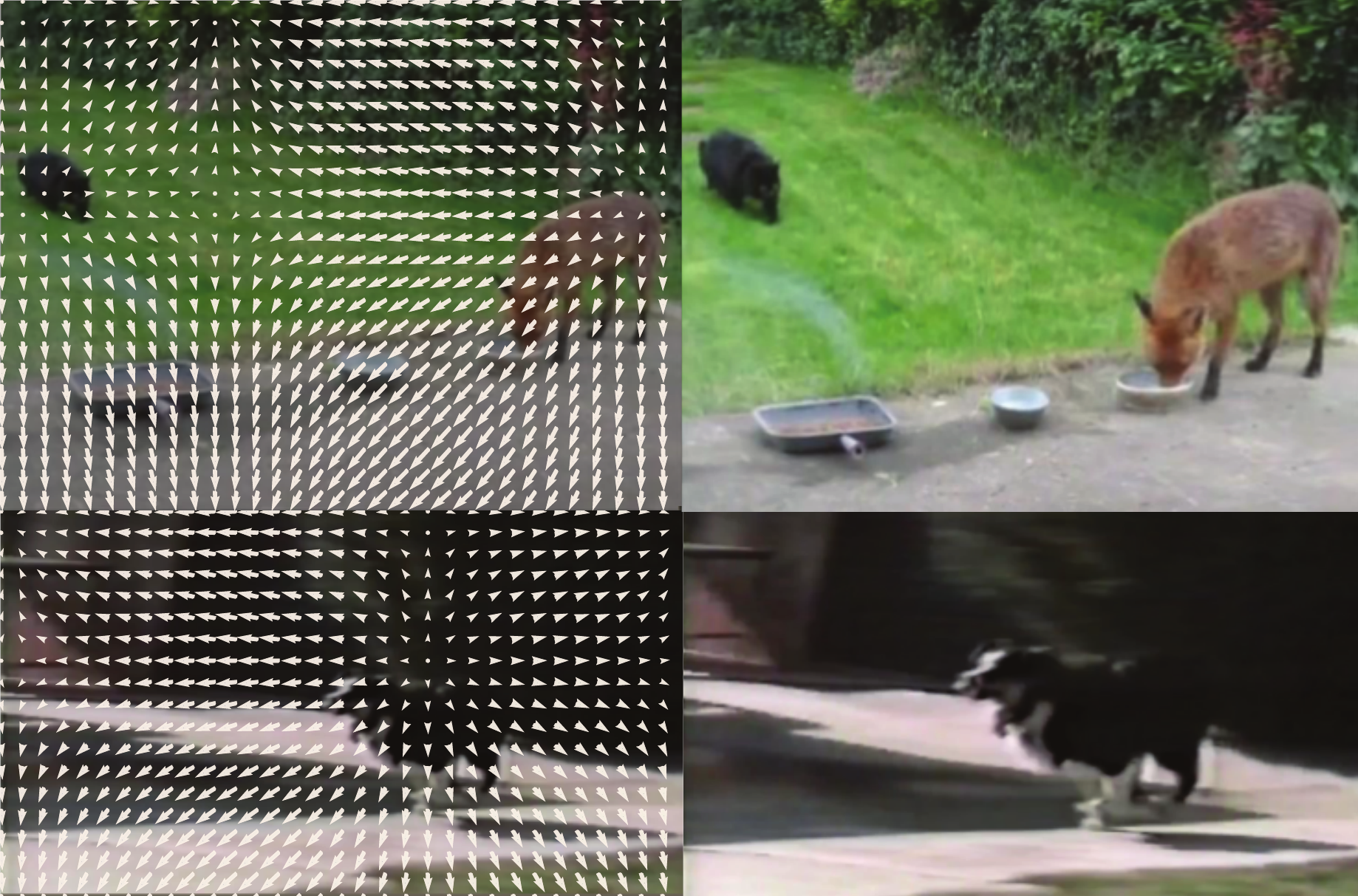}
\caption{{\bf Deformation field of TPS transformation}. Left shows the deformation field overlaid on original images. Right shows the resampled images.}
\label{fig:vector_field}
\vspace*{-12mm}
\end{wrapfigure}

\vspace*{5mm}
\noindent\textbf{Sampler.} In the final step, the sampler takes the sampling grid $G$, along with the input image ${\mbf f_{i+1}}$ to produce the detail-preserving resampled image ${\mbf f'_{i+1}}$. Figure~\ref{fig:vector_field} shows the deformation field obtained from $G$ and the resampling results for two example images.
\subsubsection{Regularization} \label{method:regul}
Learning the parameters of the localization network, without direct guidance on where to magnify, results in inhomogeneous distortions and may not preserve the desired detail. To address this, we propose to regularize the sampling grid $G$ through a non-parametric attention-based sampling method \cite{zheng2019looking}. This resampling method takes as input a saliency map and generates a sampling grid that preserves the salient regions in the map. We propose to use the sampling grid generated by this method as a supervision signal for our sampling module to learn an explicit zooming effect. However, despite obtaining superior sampling results compared to alternative methods \cite{recasens2018learning,li2017dynamic}, this approach~\cite{zheng2019looking} is non-differentiable, computationally expensive, and may generate undesirable deformations when multiple objects with various saliency levels appear in the same image. This method decomposes the saliency map into two marginal distributions over $x$ and $y$ axes. Unfortunately, this marginalization leads to undesirable distortions for low saliency regions located on the same row/column as an object with a higher saliency level. More concretely, if the coordinates $(i, j)$ and $(i', j')$ in the saliency map have high values, the resulting sampling grid is not only dense at $(i, j)$ and $(i', j')$, but also at $(i, j')$ and $(i', j)$ regardless of its saliency level. This error can be problematic when there are multiple objects with different saliency levels in the image. While our resampling module is fully differentiable and computationally inexpensive, getting an unmediated supervision from \cite{zheng2019looking} may carry the same undesirable artifacts to our sampler. To address this issue, we design the following weighted \textbf{$\ell_2$}-loss function:
\begin{equation}
Loss_\text{grid} = \sum_{h,w} M \odot \|G-G' \|_2,
\end{equation}

where $G$ is a grid generated by our resampling module, $G'$ is the grid generated by the attention-based sampling method \cite{zheng2019looking}, and $M$ is a weighted mask with the spatial dimension of $h\times w$. The weighted mask gets assigned different values for the small objects ($O_s$), large objects ($O_l$), and background ($bg$). Categorising the objects as small or large is based on the area parameter $\alpha$. If the saliency map generated in Step~\ref{sal_gen} only contains small objects or only large objects we set $(O_s, O_l, bg)$ to $(1, 0, \gamma)$ and $(0, 1, \gamma)$, respectively. Otherwise if it contains both small and large objects to $(1, 0, 0)$. Intuitively, when the saliency map is composed of a single saliency level (e.g., multiple small objects), \cite{zheng2019looking} generates plausible zooming effects for all the objects and, therefore, we can get full supervision for the entire grid. In contrast, when the saliency map is composed of multiple saliency levels (e.g., a combination of small and large objects), the method \cite{zheng2019looking} may distort objects with lower saliency. Therefore, we choose not to get supervision in those regions by masking them to zero. Note that having a down-weighted supervision (soft) in these regions did not lead to any improvements.

\begin{figure*}[t]
\centering
\includegraphics[width=1.0\linewidth]{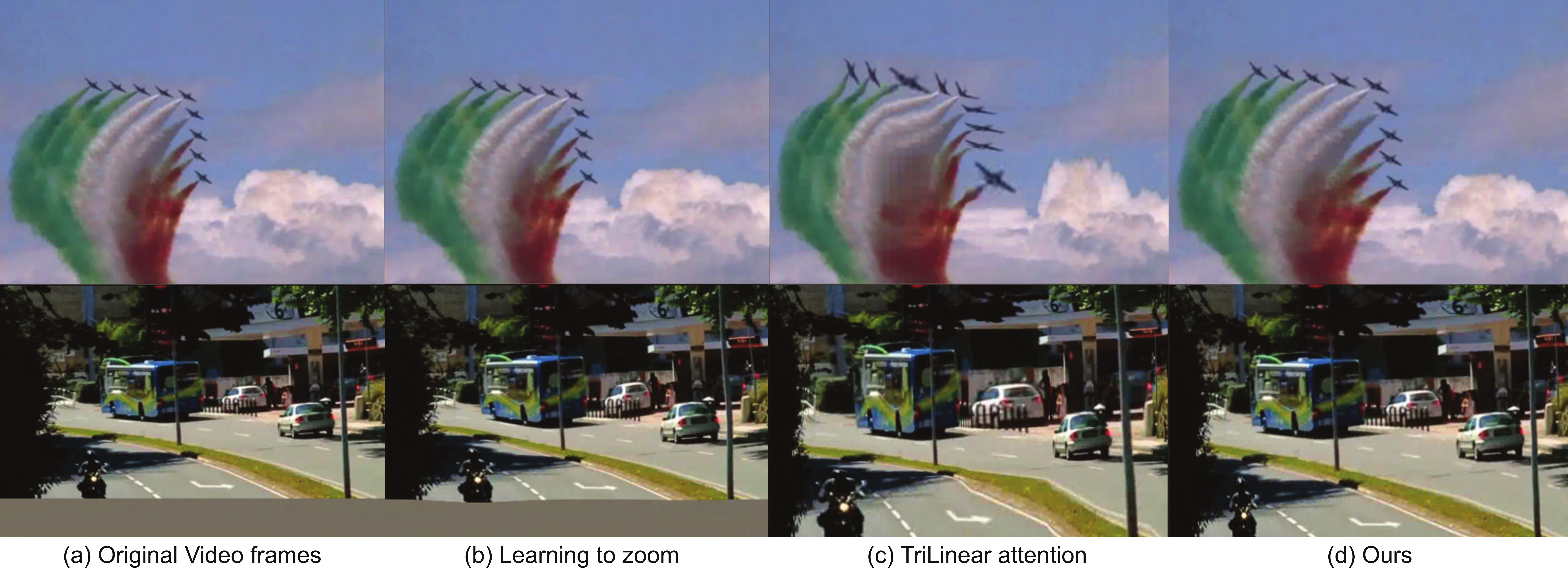}
\caption{{\bf Comparison of different input resampling methods}. (a) shows example video frames from \vid~dataset. (b), (c), and (d) show the result of resampling using learning to zoom~\cite{recasens2018learning}, TriLinear attention \cite{zheng2019looking}, and our proposed resampling module, respectively. Our resampling module effectively preserves the spatial resolution of salient objects and despite being regularized by \cite{zheng2019looking}, does not generate artifacts in background regions. This is evident from the resampled images (see the flying jets and the median barrier separating the cars).}
\label{fig:qualitative_sampling}
\vspace*{-5mm}
\end{figure*}

We train our network end-to-end by adding $Loss_\text{grid}$ to the detection loss. As can be seen in Figure~\ref{fig:qualitative_sampling}, our resampling module generally generates similar zooming effects to \cite{zheng2019looking} yet largely eliminates its distortions (see the flying jets and the median barrier separating the cars).


\subsection{Inverse transformation module}
Given the bounding box outputs of the detector $D$ for a resampled image, we apply an inverse transformation to bring the bounding boxes coordinates back to the original image grid. This is achieved by subtracting the grid displacement offset from the bounding box coordinates. As the bounding box coordinates are floating point values, for each bounding box coordinate, we obtain the exact original coordinate by linearly interpolating the displacements corresponding to its two closest cells on the deformation grid.


\section{Experiments}
\label{sec:exp}
To demonstrate the efficacy of \methodname, we conduct experiments on two large-scale video object detection datasets ImageNet-VID~\cite{russakovsky2015imagenet} and UA-DETRAC~\cite{lyu2017ua,lyu2018ua,CVIU_UA-DETRAC} as described in Section~\ref{sec:exp:datasets}. We provide comparisons to state-of-the-art video object detection models and demonstrate that \methodname~outperforms the state of the art while significantly reducing computational costs in Section~\ref{sec:exp:results}. Additionally, to demonstrate the efficacy of our regularized sampling module, we compare our method with other competing sampling approaches. Finally, we present several ablation studies to discuss the effect of several design choices on the performance of our method in Section~\ref{sec:exp:ablation}.



\subsection{Experimental setup}
\label{sec:exp:setup}

\textbf{Datasets.}
\label{sec:exp:datasets}
We evaluate our method on two large video object detection datasets: ImageNet-VID~\cite{russakovsky2015imagenet} and UA-DETRAC~\cite{lyu2017ua,lyu2018ua,CVIU_UA-DETRAC}. ImageNet-VID contains $30$ object categories with $3862$ training and $555$ validation videos. Following the protocols in~\cite{liu2019looking,zhu2017deep}, during training, we also use a subset of ImageNet-DET training images, which contain the same 30 categories. We report standard mean average precision (mAP) at IoU=$0.5$ on the validation set, similar to~\cite{liu2019looking,zhu2017deep}. UA-DETRAC consists of $10$ hours of video (about $140k$ frames in total) captured from $100$ real-world traffic scenes. The scenes include urban highways, traffic crossings, T-junctions, etc., and the bounding box annotations are provided for vehicles. The dataset comes with a partitioning of $60$ and $40$ videos as train and test data, respectively. Following~\cite{habibian2021skip}, average precision (AP), averaged over multiple IoU thresholds varying from $0.5$ to $0.95$ with a step size of $0.05$ is reported on the test data.

\medskip\noindent\textbf{Implementation details.} 
\label{sec:exp:implemetation_detail}
We use different variants of EfficientDet \cite{tan2020efficientdet}, namely D0-D4, as detectors in our video object detection framework. \methodname~has two separate object detectors, one for the key frame ($\mathscr{D}_\text{key}$) and another for all succeeding frames ($\mathscr{D}$). In our experiments, we use two successive scaled-up variants of EfficientDet, for example, EfficientDet-D3 and EfficientDet-D2, where the heavier model is applied to the key frame and the lighter one to the rest of the frames. In this particular example, we refer to our model as \methodname~with EfficientDet-D2. We follow the same procedure, for baseline EfficientDet models without resampling. 

We first trained the resampling module independently from the detection network using the regularization loss described in \ref{method:regul}. For both datasets, we then trained the EfficientDet networks, pre-trained on MS-COCO~\cite{lin2014microsoft}, in an image-based fashion. In the final step, we fine-tuned the resampling module and the object detection networks end-to-end. The complete details for training are provided in the Appendix~\ref{app:train}.
 
During inference, key frames are picked once every $S$ frames ($S = 2 \sim 32$ frames) and passed to $\mathscr{D}_\text{key}$ while the succeeding $S-1$ frames are processed by $\mathscr{D}$. For \vid~and \detrac~experiments, we set $S$ to $16$ and $32$, respectively. We set the parameter of the saliency map generator $\tau$ to $0.5$. We set $\gamma$ controlling the weight of the regularizer in background regions to $0.5$. We report the average per-frame computation cost of our model by considering the FLOPs of $\mathscr{D}_\text{key}$, $\mathscr{D}$, and resampling module. In our experiments, unless otherwise specified, for both baseline models and \methodname, predictions are made for odd frames $f_i$ where $i\in\{1,3,5,...,N-1\}$ and propagated to the next frame $f_{i+1}$ without further processing. For \methodname, this means propagating the saliency maps every other frame. For both baseline and \methodname,~this setup yields up to $50\%$ reduction in FLOPs with only a small drop in the accuracy. To achieve the highest performance on each benchmark, we still apply our model densely to all frames and explicitly mention dense prediction if that is the case.



\subsection{Results}
\label{sec:exp:results}

\begin{figure}[t]
\centering
\includegraphics[width=0.9\linewidth]{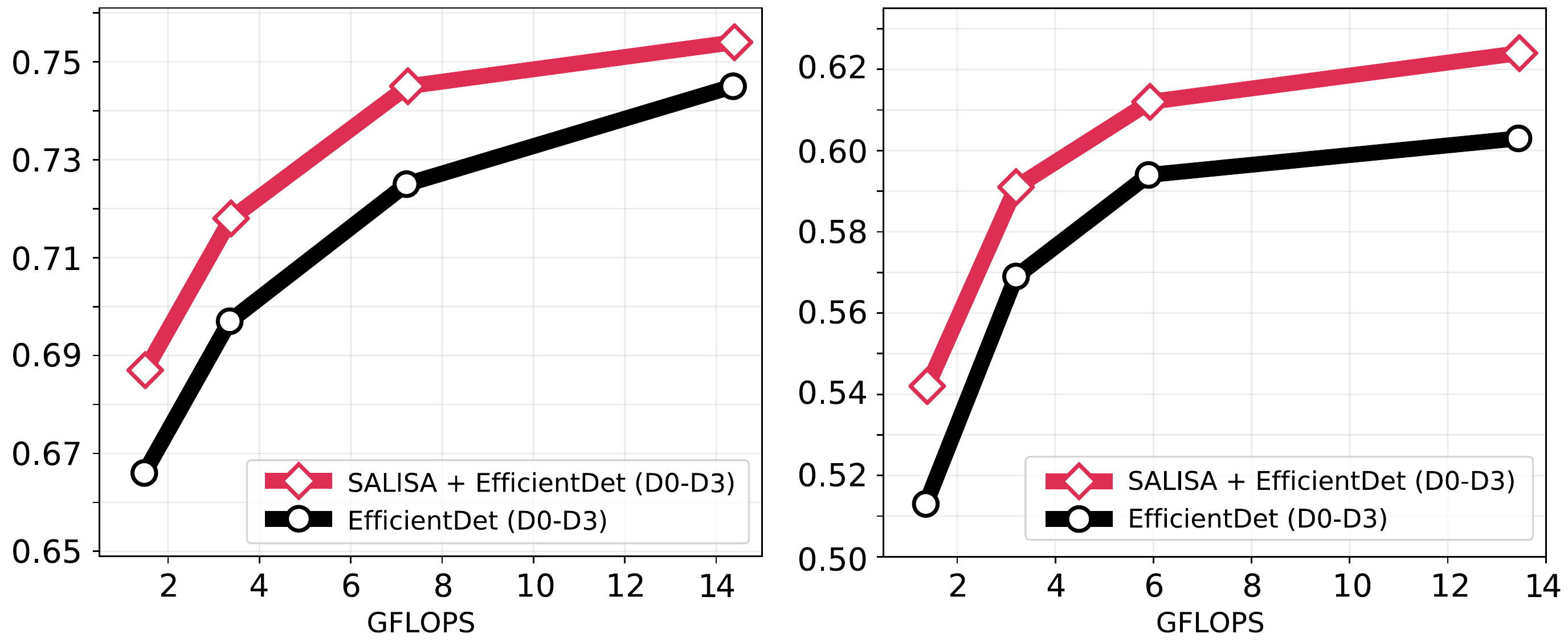}
\caption{Performance comparison of baseline EfficientDet~\cite{tan2020efficientdet} and corresponding \methodname+EfficientDet models on ImageNet-VID (left) and UA-DETRAC (Right).}
\label{fig:curve_both}
\vspace*{-5mm}
\end{figure}

\noindent\textbf{Comparison to state of the art: \detrac.} We compare~\methodname~to several image and video object detectors on the \detrac~dataset: EfficientDet~\cite{tan2020efficientdet} as the state of the art in efficient object detection in images and the main baseline for \methodname, Deep Feature Flow (DFF)~\cite{zhu2017deep} as  a  seminal  work  on  efficient  object  detection, 
and SpotNet~\cite{perreault2020spotnet} as the highest performing method on the UA-DETRAC benchmark.
Figure~\ref{fig:curve_both} presents accuracy  vs. computations  trade-off  curves  for \methodname~and the baseline EfficientDet models (D0-D3) for video object detection in \detrac. As can be seen, our method consistently outperforms the baseline EfficientDet models. Importantly, \methodname~with EfficientDet-D2 ($61.2\%$) outperforms EfficientDet-D3 model ($60.3\%$) at lower than half the computational cost. In the low-compute regime, \methodname~with EfficientDet-D0 outperforms the baseline EfficientDet-D0 model by $2.9\%$. The comparison with competing methods is shown in Table~\ref{table:sota_detrac}. We outperform DFF \cite{zhu2017deep} both in terms of computational costs and accuracy. When densely applied to all frames, \methodname~with EfficientDet-D3 achieves state-of-the-art mAP of $62.9\%$ on \detrac~at a much lower computational cost than SpotNet (972 VS. 40 GFlops).

\begin{table}[t]
\caption{Comparison with state of the art on \detrac.}
\label{table:sota_detrac}
\centering
\resizebox{0.6\columnwidth}{!}
{
\begin{tabular}{l|c|c|c}
\hline
\cellcolor{mygray} \textbf{Method} & \cellcolor{mygray} \textbf{Backbone} & \cellcolor{mygray} \textbf{mAP (\%)} & \cellcolor{mygray} \textbf{FLOPs (G)} \\
\hline

DFF~\cite{zhu2017deep} & ResNet-50 & 52.6 & 75.3 \\

SpotNet~\cite{perreault2020spotnet} &  CenterNet \cite{zhou2019objects} & 62.8 & 972.0 \\
EfficientDet~\cite{tan2020efficientdet} & EfficientNet-B2 & 59.4 & 5.9 \\
EfficientDet~\cite{tan2020efficientdet} & EfficientNet-B3 & 60.3 & 13.4 \\
\textbf{\methodname(Ours)} & EfficientNet-B2 & 61.2 & 5.9\\
\textbf{\methodname(Ours)} & EfficientNet-B3 & 62.4 & 13.4 \\
\hline
EfficientDet~\cite{tan2020efficientdet} & EfficientNet-B0 & 51.3 & 1.36 \\
EfficientDet~\cite{tan2020efficientdet} & EfficientNet-B1 & 56.9 & 3.20 \\

\textbf{\methodname(Ours)} & EfficientNet-B0 & 54.2 & 1.39 \\
\textbf{\methodname(Ours)} & EfficientNet-B1 & 59.1 & 3.23\\

\hline
\end{tabular}
}
\end{table}

\begin{table}[t]
\caption{Comparison with state of the art on ImageNet-VID. $^*$ indicates that the model has been applied every three frames.}
\label{table:sota_imvid}
\centering
\resizebox{0.7\columnwidth}{!}
{
\begin{tabular}{l|l|c|c}
\hline
\cellcolor{mygray} \textbf{Method} & \cellcolor{mygray} \textbf{Backbone} & \cellcolor{mygray} \textbf{mAP (\%)} & \cellcolor{mygray} \textbf{FLOPs (G)} \\
\hline
DFF (R-FCN)~\cite{zhu2017deep} & ResNet-101 & 72.5 & 34.9\\
PatchNet (R-FCN)~\cite{mao2021patchnet} &  ResNet-101 & 73.1 & 34.2 \\
TSM \cite{lin2019tsm} & ResNext101~\cite{xie2017aggregated} & 76.3 & 169 \\

SkipConv~\cite{habibian2021skip} & EfficientNet-B2 & 72.3 & 9.2 \\
SkipConv~\cite{habibian2021skip} & EfficientNet-B3 & 75.2 & 22.4 \\
EfficientDet~\cite{tan2020efficientdet} & EfficientNet-B2 & 72.5 & 7.2 \\
EfficientDet~\cite{tan2020efficientdet} & EfficientNet-B3 & 74.5 & 14.4 \\
\textbf{\methodname(Ours)} & EfficientNet-B2 & 74.5 & 7.2 \\
\textbf{\methodname(Ours)} & EfficientNet-B3 & 75.4 & 14.4\\
\hline

Mobile-SSD & MobileNet-V2 & 54.7 & 2.0 \\
PatchWork~\cite{chai2019patchwork} & MobileNet-V2 & 57.4 & 0.97 \\
PatchNet (EfficientDet)~\cite{mao2021patchnet} & EfficientNet-B0 & 58.9 & 0.73\\
Mobile-DFF~\cite{zhu2017deep} & MobileNet & 62.8 & 0.71\\
TAFM (SSDLite)~\cite{liu2018mobile}& MobileNet-V2 & 64.1 & 1.18 \\
SkipConv~\cite{habibian2021skip} & EfficientNet-B0 & 66.2 & 0.98 \\
SkipConv~\cite{habibian2021skip} & EfficientNet-B1 & 70.5 & 2.90 \\
EfficientDet~\cite{tan2020efficientdet} & EfficientNet-B0 & 66.6 & 1.48 \\
EfficientDet~\cite{tan2020efficientdet} & EfficientNet-B1 & 69.7 & 3.35 \\
\textbf{\methodname(Ours)} & EfficientNet-B0$^{*}$ & 67.4 & 0.86 \\
\textbf{\methodname(Ours)} & EfficientNet-B0 & 68.7 & 1.50 \\
\textbf{\methodname(Ours)} & EfficientNet-B1 & 71.8 & 3.38 \\
\hline

\end{tabular}
}
\end{table}

\begin{figure*}[t]
\centering
\includegraphics[width=1\linewidth]{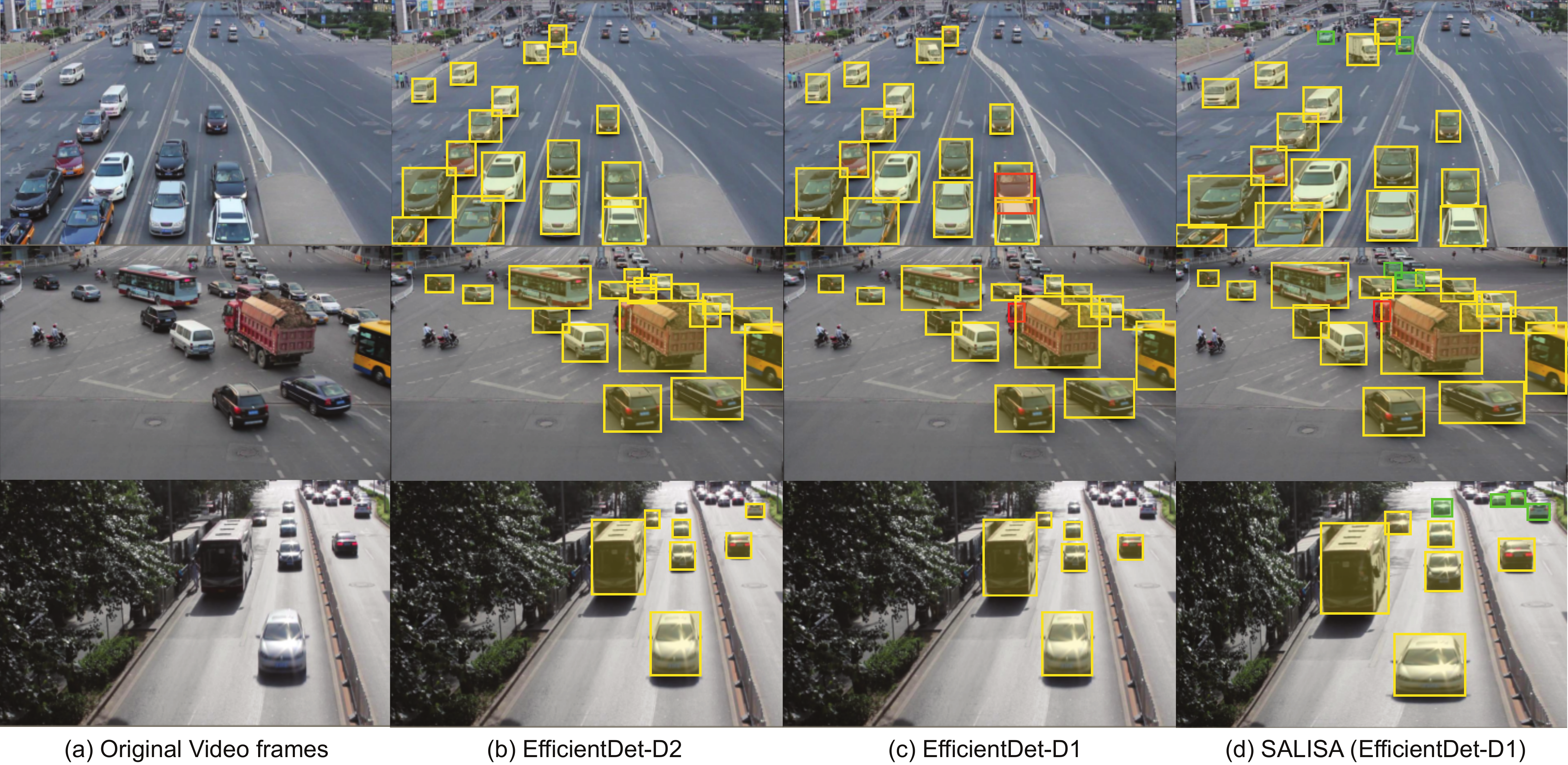}
\caption{{\bf Detection results on the \detrac~test set}. Yellow boxes indicate true detections, red indicates false positive detections, and green boxes refer to new detections produced by our method as a result of input resampling. (a) shows the original video frames from \detrac~dataset, (b), (c), and (d) show detection results generated by EfficientDet-D2 ($\mathscr{D}_{key}$), the baseline EfficientDet-D1 detector, and \methodname~with EfficientDet-D1, respectively. As can be seen from the detections in the first and third row of \textit{(c)}, the right side of the road in the first image and the vegetation in the third image have been pushed to the side to enable magnifying salient objects. This detail-preserving down-sampling has allowed for discovery of new objects that were otherwise missed by the baseline object detector.}
\label{fig:detectiono_visuals}
\vspace*{-5mm}
\end{figure*}

Finally, the results presented in Figure~\ref{fig:detectiono_visuals} show some challenging object detection scenes from the test set. For example, the frame in the first row shows a crowded scene with many vehicles which makes zooming particularly challenging. Our model has squeezed the right side of the road to increase the resolution of the salient objects. This has enabled \methodname~with EfficientDet-D1 to detect new cars which were neither detected by the baseline EfficientDet-D1 nor by the heavier keyframe detector EfficientDet-D2.



\vspace{1mm}
\noindent\textbf{Comparison to state of the art: \vid.} The experimental results of \methodname~on the ImageNet-VID dataset are presented in Table~\ref{table:sota_imvid}. We compare our method to PatchNet~\cite{mao2021patchnet}, PatchWork~\cite{chai2019patchwork}, TSM~\cite{lin2019tsm}, DFF~\cite{zhu2017deep}, Mobile-DFF~\cite{zhu2017deep}, Mobile-SSD, TAFM~\cite{liu2018mobile}, SkipConv~\cite{habibian2021skip}, and finally EfficientDet~\cite{tan2020efficientdet} as our baseline. DFF and Mobile-DFF are flow-based methods, PatchNet is a tracking method, TAFM is an LSTM-based recurrent method, PatchWork and SkipConv conditionally limit feature computation, and TSM reduces the computation by shifting features across time. In Table~\ref{table:sota_imvid}, these methods are categorized as low compute and extremely low compute. The results show that among the extremely low compute methods, \methodname~with EfficientDet-D0 significantly outperforms Mobile-SSD, PatchWork~\cite{chai2019patchwork}, TAFM~\cite{liu2018mobile}, and SkipConv~\cite{habibian2021skip} at a lower computational cost. While PatchNet~\cite{mao2021patchnet} and Mobile-DFF~\cite{zhu2017deep} have roughly 0.15 lower GFLOPs than our lightest model, they show a significant drop in mAP ($\sim10\%$) compared to \methodname. In general, template matching techniques offer significant computational saving without improving accuracy while \methodname~can provide gains in both aspects. See Appendix~\ref{app:tracking} for additional comparison with tracking baselines.

Among the low compute methods, \methodname~with EfficientDet-D3 outperforms DFF and PatchNet by $2.9\%$ and $2.3\%$, respectively at roughly $40\%$ of their computational cost. TSM is the highest performing competitor that achieves an mAP of $76.3\%$ at the cost of $169$ GFLOPs. When densely applied to all frames, \methodname~with EfficientDet-D3, obtains an mAP of $76.4\%$ at $40$ GFLOPs. 
Figure~\ref{fig:curve_both} presents accuracy vs. computations trade-off curves  for \methodname~and EfficientDet baseline models (D0-D3) for video object detection in \vid. \methodname~consistently boosts the performance of EfficientDet variants by adaptively resampling the input. Finally, \methodname~with EfficientDet-D2 matches the mAP of the baseline EfficientDet-D3 at half the computational cost.

\vspace{2mm}
\noindent\textbf{Performance across different object sizes.}
To demonstrate the efficacy of \methodname~for detecting small objects, we report the mAP scores for different object sizes using the COCO framework~\cite{lin2014microsoft}. As shown in Table~\ref{table:object_size}, \methodname~significantly improves small object detection compared to the baseline EfficientDet models. In particular, \methodname~with EfficientDet-D1, improves the accuracy of small object detection by $77\%$ ($8.4\%$ to $14.9\%$). Surprisingly, this is even higher than $12.5\%$ mAP of EfficientDet-D3 baseline for small objects.

\noindent
\begin{minipage}[c]{0.45\textwidth}

\centering
\footnotesize
\captionof{table}{Performance comparison (mAP) across different object sizes on \detrac.} 
\label{table:object_size}
\begin{tabular}{l|ccc}
\hline

\cellcolor{mygray} \textbf{Model} & \cellcolor{mygray}  \textbf{Small} & \cellcolor{mygray}  \textbf{Medium} & \cellcolor{mygray} \textbf{Large} \\
 \hline
 EfficientDet-D0    & 6.4 & 47.8 & 72.1 \\
 EfficientDet-D1   & 8.4 & 54.8 & 75.7 \\
 EfficientDet-D2    & 12.3 & 58.1 & 77.1 \\
 EfficientDet-D3    & 12.5 & 59.1 & 78.5 \\
 \hline
 \cellcolor{mygray} \methodname-D0     & \cellcolor{mygray} 7.4 & \cellcolor{mygray} 53.1 & \cellcolor{mygray} 73.0 \\ 
 \cellcolor{mygray} \methodname-D1     & \cellcolor{mygray} 14.9 & \cellcolor{mygray} 58.2 & \cellcolor{mygray} 76.6 \\ 
 \cellcolor{mygray} \methodname-D2     &  \cellcolor{mygray} 15.3 &  \cellcolor{mygray} 59.4 &  \cellcolor{mygray} 77.7 \\
 \cellcolor{mygray} \methodname-D3     &  \cellcolor{mygray} 16.6 &  \cellcolor{mygray} 60.1 &  \cellcolor{mygray} 78.0 \\ 
 \hline
\end{tabular}

\end{minipage}
\hspace{8mm}
\begin{minipage}[c]{0.45\textwidth}
\centering

\footnotesize
\captionof{table}{Impact of input sampling method on \detrac~(upper part) and \vid~(bottom part).}
\label{table:saliency_comparison}
\centering
\begin{tabular}{l|ccc}
\hline

\cellcolor{mygray} \textbf{Resampling method} & \cellcolor{mygray}  \textbf{D0} & \cellcolor{mygray}  \textbf{D1} & \cellcolor{mygray} \textbf{D2} \\
 \hline
 TPS-STN \cite{jaderberg2015spatial}  & 51.7 & 57.6 & 60.8 \\
Learning to zoom \cite{recasens2018learning}          & 39.2 & 47.7 & 52.1 \\ 
Trilinear attention \cite{zheng2019looking}       & 53.6 & 58.7 & 61.5 \\
\textbf{Our resampling module}                   & \textbf{55.6} & \textbf{61.4} & \textbf{62.7} \\ 
 \hline
\cellcolor{mygray} TPS-STN \cite{jaderberg2015spatial}                   & \cellcolor{mygray} 69.1 & \cellcolor{mygray} 71.1 & \cellcolor{mygray} 73.7       \\
\cellcolor{mygray} Learning to zoom \cite{recasens2018learning}          & \cellcolor{mygray} 69.0 & \cellcolor{mygray} 71.3 & \cellcolor{mygray} 74.9 \\ 
\cellcolor{mygray} Trilinear attention \cite{zheng2019looking}      & \cellcolor{mygray} 69.4 & \cellcolor{mygray} 71.5 & \cellcolor{mygray} 74.8 \\
\cellcolor{mygray} \textbf{Our resampling module}                      & \cellcolor{mygray} \textbf{69.7} & \cellcolor{mygray} \textbf{72.4} & \cellcolor{mygray} \textbf{75.2} \\ 
 \hline
\end{tabular}

\end{minipage}
\vspace{2mm}

The mAP of medium-sized object detection increases from $54.8$ to $58.2$. There is no significant change in the performance of large object sizes as the base model can also effectively detect them. That is why lighter models with smaller inputs benefit more compared to heavier models that already receive a high resolution input.


\vspace{3mm}
\noindent\textbf{Comparison to different sampling approaches.} In this experiment, we compare the performance of various sampling approaches \cite{jaderberg2015spatial,recasens2018learning,zheng2019looking} for a detail-preserving downsampling on both \vid~ and \detrac~datasets.
To this end, we substitute our resampling module with these methods and use the same training protocol discussed in \ref{sec:exp:implemetation_detail}.
The results are presented in Table~\ref{table:saliency_comparison}. We first compare our resampling module to TPS-STN \cite{jaderberg2015spatial}. As can be seen, our regularization scheme is crucial for improving the results. TPS-STN without our regularizer, barely improves upon the baseline EfficientDet models. Our resampling module also yields a higher accuracy compared to \cite{recasens2018learning} and \cite{zheng2019looking} on both \vid~and \detrac~datasets. While the gap in performance in different resampling methods is small on the \vid~dataset, \methodname~greatly benefits from our resampling module on \detrac~with a gap of more than $2\%$ mAP. The videos in the \vid~dataset are mostly comprising one or two objects. The videos in the  \detrac~dataset, in contrast, include mostly crowded scenes with many objects in each frame. We conjecture that, in such multi-object wild videos the undesirable deformations induced by \cite{recasens2018learning} and \cite{zheng2019looking} can lower their benefits.
Overall, our resampling module consistently outperforms competitors in all settings.

\vspace{3mm}
\noindent\textbf{Wall-clock timing} We report the wall-clock timing (msec) of \methodname~ and the baseline EfficientDet models using Nvidia Tesla-V100 32GB. The inference time of EfficientDet and \methodname~for a batch size of one are as follows: \textbf{D0:} 49.4 vs 50.2, \textbf{D1:} 91.0 vs 95.4, \textbf{D2:} 152.7 vs 159.4, and \textbf{D3:} 304.8 vs 313.4. The overhead of our sampler ($0.06$ GFLOPs) is very small primarily because of its small input size.


\subsection{Ablation study}
\label{sec:exp:ablation}
\noindent\textbf{Analysis of the area threshold.} 
The resampling module preserves the resolution of all salient objects, however, it gives extra focus in preserving the resolution of small object. The area threshold $\alpha$ determines which objects should be considered as small in the saliency map. As seen in Table~\ref{table:area_threshold}, generally, a smaller value of $\alpha$ improves small object detection without hurting the detection performance of medium and large objects. This is because smaller $\alpha$ values increasingly move the focus of the sampler to preserve the resolution of smaller objects. Extremely small $\alpha$ values (\eg~$0.1\%$), however, may lead to reduced small object detection performance as just limited number of small objects receive extra focus.

\begin{figure}[t]
\centering
\includegraphics[width=0.98\linewidth]{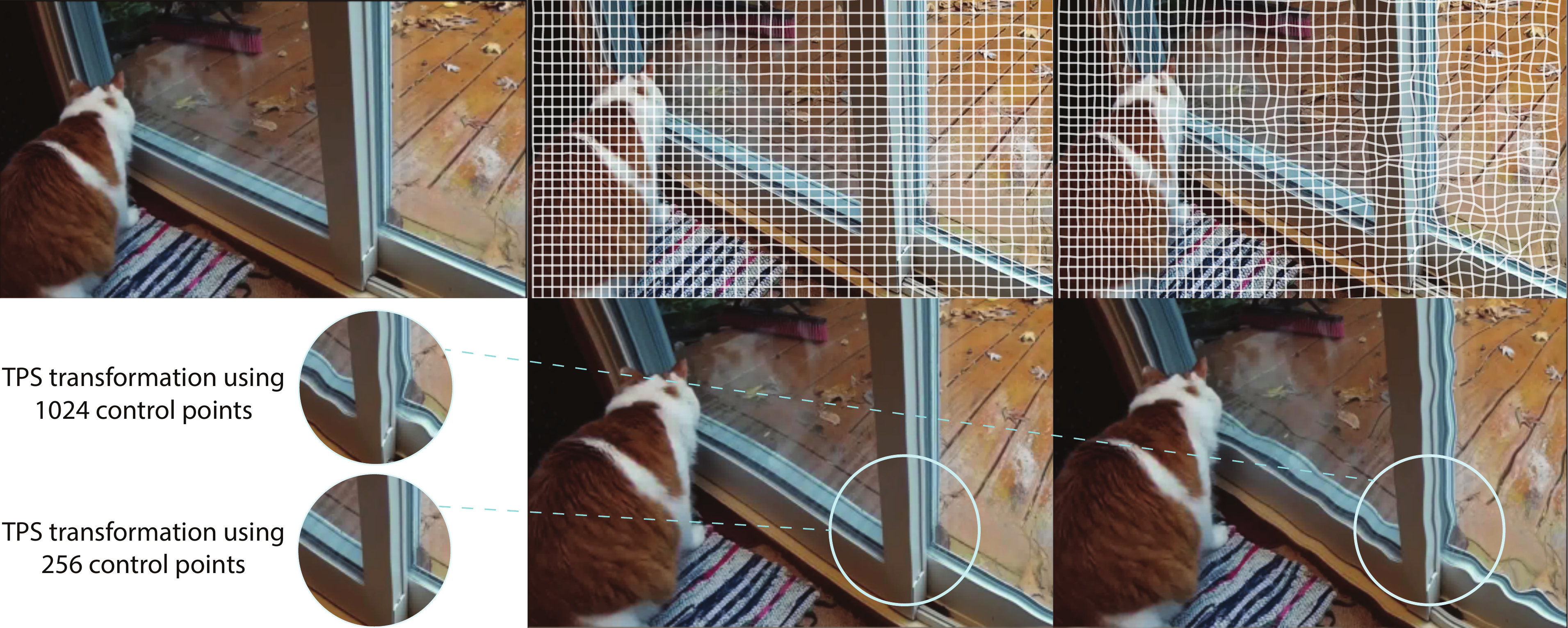}
\caption{{\bf Effect of the number of control points on TPS transformation.} The top row shows the grid deformations produced by TPS with $256$ (middle column) and $1024$ control points (right column). The bottom row shows the corresponding resampled images.}
\label{fig:tps_points}
\vspace*{-5mm}
\end{figure}

\vspace{3mm}
\noindent\textbf{Number of control points in TPS.} Estimating the parameters of TPS relies on defining correspondences between a set of control points and their displacements. Increasing the number of control points generally increases the flexibility of TPS. While we observe a reduction in $loss_{grid}$ when we increase the number of control points from $256$ to $1024$, we also notice more fluctuations and artifacts in the resulting resampled images as shown in Figure~\ref{fig:tps_points}. By increasing the number of control points from $256$ to $1024$, the mAP of \methodname~with EfficientDet-D0 on \detrac~drops from $54.2$ to $45.7$, and for EfficientDet-D1 from $59.1$ to $49.5$.
As defining $256$ control points gives better detection results, we set the number of TPS control points to $256$.

\vspace{3mm}
\noindent\textbf{Combinations of heavy-light networks.} In this ablation, we analyze different combination possibilities for the keyframe detector ($\mathscr{D}_{key}$) and the main detector ($\mathscr{D}$), for example combining EfficientDet-D3 and EfficientDet-D0. In general, we observe that the additional costs of heavier networks undermine the extra gained accuracy. For example, on \detrac, using \methodname~with D3 as  $\mathscr{D}_{key}$ and D0 as $\mathscr{D}$ increases the overall mAP just by 0.3\%, compared to D1-D0 combination, while increasing the FLOPS from $1.39G$ to $1.98G$. Check Appendix~\ref{app:combo} for the results of all combinations.

\begin{table}[t]
\footnotesize
\caption{Effect of $\alpha$ on the mAP of \methodname~with EfficientDet-D1.}
\label{table:area_threshold}
\centering

\begin{tabular}{l|ccc}
\hline
\cellcolor{mygray} \textbf{Threshold} & \cellcolor{mygray} \textbf{Small} & \cellcolor{mygray} \textbf{Medium} & \cellcolor{mygray} \textbf{Large} \\
 \hline
$\alpha=2.0$ & 13.1 & 58.1 & 76.3 \\ 
$\alpha=1.0$ & 13.3 & 58.2 & 76.6 \\
$\alpha=0.5$ & \textbf{14.9} & 58.2 & 76.6 \\ 
$\alpha=0.1$ & 13.4 & 58.1 & 76.9 \\ 

\hline
\end{tabular}

\end{table}



\section{Discussion and Conclusion}
In this paper, we proposed \methodname, a saliency-based input sampling technique for efficient video object detection. \methodname~performs a nonuniform downsampling of the input by retaining the fine-grained details of a high-resolution image while allowing for heavy downsampling of background areas. The resulting image is spatially smaller, leading to a reduction in computation costs, but preserves the important details enabling a performance comparable to a high-resolution input. We propose a novel and fully differentiable resampling module based on thin plate spline spatial transformers that generates artifact-free resampled images.
\methodname~achieves state-of-the-art accuracy on the \vid~and \detrac~video object detection datasets in the low compute regime. In particular, it offers significant improvements in the detection of small- and medium-sized objects. A limitation of our model is that, it preserves high-resolution details by downsampling background regions more aggressively. However, when the scene is fully covered with objects, e.g. in a heavy traffic scene, proper zooming is less achievable as there is less background pixels to sub-sample.



\clearpage
%
%
\bibliographystyle{splncs04}
\bibliography{0_main}

\clearpage
\appendix
\section*{Appendix}
\section{Thin-plate Spline transformation}
\label{app:TPS}
In this section, we explain the mathematical details of the Thin Plate Spline (TPS) warping function. Given a set of $N$ control points on a 2D grid $\overset{.}{P}\in \mathbb{R}^{N\times 2}$, and their transformed positions $\overset{.}{V}\in \mathbb{R}^{N\times 2}$, we solve for two functions $f_{x'}$ and $f_{y'}$, from which we can sample discrete displacements along x and y coordinates, as follows:

\begin{equation}
f_{x'}(x,y)=a_1 +a_2 x + a_3 y +\displaystyle\sum\limits_{i=0}^N \alpha_i U(\lVert(x_i,y_i)-(x,y)\rVert)
\end{equation}
\begin{equation}
f_{y'}(x,y)=a_4 +a_5 x + a_6 y +\displaystyle\sum\limits_{i=0}^N \beta_i U(\lVert(x_i,y_i)-(x,y)\rVert)
\end{equation}

The term $r_{ij}=\lVert(x_i,y_i)-(x_j,y_j)\rVert$ represents the distance between the control points $\overset{.}{P}_j=(x_j,y_j)$ and $\overset{.}{P}_i=(x_i,y_i)$, and $U(r)=r^2log(r)$ is the radial basis kernel. The six parameters $a_1,a_2,...,a_6$ correspond to the global affine transformation of TPS, and $2N$ parameters $(\alpha_i, \beta_i) \in \{1, 2, ..., N\}$ correspond to local transformation. Let us define the matrices $K$, $P$, $W$, and $V$ as follows:

\begin{equation}
K=
\begin{bmatrix} 
0 & U(r_{12}) & ... & U(r_{1N}) \\
U(r_{21}) & 0 & ... & U(r_{2N}) \\
... & ... & ... & ... \\
U(r_{N1}) & U(r_{N2}) & ... & 0 \\
\end{bmatrix}, N\times N, 
\quad\quad
P=
\begin{bmatrix} 
1 & x_1 & y_1 \\
1 & x_1 & y_1 \\
... & ... & ...  \\
1 & x_N & y_N \\
\end{bmatrix}, N\times 3,
\end{equation}

\begin{equation}
W=
\begin{bmatrix} 
\alpha_1 & \beta_1 \\
\alpha_1 & \beta_1 \\
 ... & ...  \\
\alpha_N & \beta_N \\
a_1 & a_4 \\
a_2 & a_5 \\
a_3 & a_6 \\
\end{bmatrix}, (N+3)\times 2, 
\quad\quad
V=
\begin{bmatrix} 
x'_1 & y'_1 \\
x'_1 & y'_1 \\
 ... & ...  \\
x'_N & y'_N \\
0 & 0 \\
0 & 0 \\
0 & 0 \\
\end{bmatrix}, (N+3)\times 2,
\end{equation}

The TPS coefficients can be calculated by solving the following linear problem:

\begin{equation}
W
=
\underbrace{\begin{bmatrix}
    \begin{array}{c|c}
  K & P \\
  \hline
  P^T & O
    \end{array}
\end{bmatrix}}^{~~~~~~~~~~~~~~~~~~~-1}_\text{$L$}
\times
V
\end{equation}

Once $W$ is calculated, we plug it into equations (1) and (2), to sample $f_{x'}$ and $f_{y'}$ given a point $(x,y)$.

\section{Training protocol}
\label{app:train}
We first trained the resampling module independently from the detection network using the regularization loss described in Section 3.2.1 of the main paper. For this purpose, we first generated saliency maps from the ground-truth annotations of the UA-DETRAC training dataset and trained the sampler using Adam optimizer~\cite{adam} with a learning rate of $1e^{-3}$ for 10 epochs. Note that we did not pre-train our resampling module on the \vid~dataset as the pretrained weights from UA-DETRAC were already suitable for \vid~as well. For both datasets, we then trained the EfficientDet networks, pre-trained on MS-COCO~\cite{lin2014microsoft}, in an image-based fashion using SGD optimizer with momentum $0.9$, weight decay $4e^{-5}$, and an initial learning rate of $0.01$.
For ImageNet-VID, we trained the models for $7$ epochs and the learning rate was dropped with a factor of $0.1$ at epochs $3$ and $6$. For UA-DETRAC, we trained the models for $4$ epochs and dropped the learning rate with a factor of $0.1$ at epoch $3$. In the final step, we fine-tuned the resampling module and the object detection networks end-to-end using SGD with a learning rate of $1e^{-3}$ for $3$ epochs. The models were trained with a mini-batch size of $4$ using four GPUs and synchronized batch-norm. We used standard data augmentations \cite{tan2020efficientdet} commonly used to train EfficientDet in our experiments.

\section{Ablations}
\subsection{Combinations of Heavy-light networks.}
\label{app:combo}
We analyze different combination possibilities for the keyframe detector ($\mathscr{D}_{key}$) and the main detector ($\mathscr{D}$). The results are presented in Table~\ref{tab:combo} for different object size categories. As can be seen, there is no extra gain in medium- and large-sized object detection when combining the main detector with more expensive key frame detectors. The mAP of small object detection improves. However, the additional costs of heavier networks undermine the extra gained accuracy.

\subsection{Comparison to tracking baselines on \detrac} 
\label{app:tracking}
We compared SALISA with two tracking baselines in Table~\ref{table:tracking}. D1+D0 is a baseline that performs detection on the key frame with EfficientDet-D1 and uses EfficientDet-D0 for the next 7 subsequent frames (no tracking). D1+Copy is a baseline that performs detection on the key frame with EfficientDet-D1 and copies the obtained boxes for the next 7 frames. In contrast, D1+SiamFC updates the boxes in non-key frames  using SiamFC tracker \cite{bertinetto2016fully}. SALISA with the same setup (EfficientDet-D1 as the key frame detector and EfficientDet-D0 for the next 7 frames), outperforms all these methods by a large margin.

\begin{table}[htb]
    \begin{subtable}[h]{0.22\textwidth}
        \centering
        \begin{tabular}{|l|ccc|}
        \hline
        \cellcolor{mygray} \textbf{} & \cellcolor{mygray} \textbf{D0} & \cellcolor{mygray} \textbf{D1} & \cellcolor{mygray} \textbf{D2} \\
         \hline
        \cellcolor{mygray} \textbf{D1} & 7.4 & - & - \\ 
        \cellcolor{mygray} \textbf{D2} & 7.4 & 14.9 & - \\
        \cellcolor{mygray} \textbf{D3} & 7.8 & 15.3 & 15.3 \\ 
        
        \hline
        \end{tabular}
       \caption{Small (mAP)}
    \end{subtable}
    \hfill
    \begin{subtable}[h]{0.22\textwidth}
        \centering
        \begin{tabular}{|l|ccc|}
        \hline
        \cellcolor{mygray} \textbf{} & \cellcolor{mygray} \textbf{D0} & \cellcolor{mygray} \textbf{D1} & \cellcolor{mygray} \textbf{D2} \\
         \hline
        \cellcolor{mygray} \textbf{D1} & 53.1 & - & - \\ 
        \cellcolor{mygray} \textbf{D2} & 53.2 & 58.2 & - \\
        \cellcolor{mygray} \textbf{D3} & 53.2 & 58.2 & 59.4 \\ 
        \hline
        \end{tabular}
       \caption{Medium (mAP)}
    \end{subtable}
    \hfill
    \begin{subtable}[h]{0.22\textwidth}
        \centering
        \begin{tabular}{|l|ccc|}
        \hline
        \cellcolor{mygray} \textbf{} & \cellcolor{mygray} \textbf{D0} & \cellcolor{mygray} \textbf{D1} & \cellcolor{mygray} \textbf{D2} \\
         \hline
        \cellcolor{mygray} \textbf{D1} & 73.0 & - & - \\ 
        \cellcolor{mygray} \textbf{D2} & 73.0 & 76.6 & - \\
        \cellcolor{mygray} \textbf{D3} & 73.0 & 76.6 & 77.7 \\ 
        \hline
        \end{tabular}
       \caption{Large (mAP)}
    \end{subtable}
    \hfill
    \begin{subtable}[h]{0.22\textwidth}
        \centering
        \begin{tabular}{|l|ccc|}
        \hline
        \cellcolor{mygray} \textbf{} & \cellcolor{mygray} \textbf{D0} & \cellcolor{mygray} \textbf{D1} & \cellcolor{mygray} \textbf{D2} \\
         \hline
        \cellcolor{mygray} \textbf{D1} & 1.39 & - & - \\ 
        \cellcolor{mygray} \textbf{D2} & 1.54 & 3.23 & - \\
        \cellcolor{mygray} \textbf{D3} & 1.98 & 3.67 & 5.96 \\ 
        \hline
        \end{tabular}
        \caption{GFLOPs}
     \end{subtable}
     \caption{Various combinations of keyframe detector $\mathscr{D}_{key}$ (rows) and the main detector $\mathscr{D}$ (columns). (a-c) shows the mAP of various combinations for small, medium, and large object detection, respectively. (d) shows the computational costs in GFLOPs.}
     \label{tab:combo}
\end{table}

\begin{table}[h]
\footnotesize
\caption{Comparison of \methodname~with several tracking baselines.}
\label{table:tracking}
\centering

\begin{tabular}{l|c|c}
\hline
\cellcolor{mygray} \textbf{Method} & \cellcolor{mygray} \textbf{mAP} & \cellcolor{mygray} \textbf{FLOPs} \\
 \hline
D1 + D0  & 52.2 & 2.9 G  \\
D1 + Copy & 30.6 & 0.8 G  \\
D1 + SiamFC \cite{bertinetto2016fully} & 51.6 & 3.1 G  \\
D1 + SALISA (D0) & 56.0 & 2.9 G \\
\hline
\end{tabular}

\end{table}

\subsection{Impact of $\tau$ for saliency map generation} 
\label{app:tau}
In this ablation, we analyze the effect of changing the parameter $\tau$ for generating the saliency maps in section $3.1$. By reducing $\tau$, we can include more uncertain detections in our saliency maps and potentially allow for a more accurate decision for those detections in subsequent frames. As can be seen in Table~\ref{table:threshold}, decreasing $\tau$ to $0.3$ gives a small but consistent improvement for all models. Decreasing $\tau$ further down to $0.1$ still comes with a small accuracy improvement compared to $\tau=0.5$. Therefore, in practice including uncertain detections can increase the performance, however, in case a model has many false positives, this could also crowd the saliency map and defeat the purpose of focused zooming. 


\begin{table}[htb]
\footnotesize
\caption{Effect of $\tau$ on the mAP of \methodname.}
\label{table:threshold}
\centering

\begin{tabular}{l|ccc}
\hline
\cellcolor{mygray} \textbf{Threshold} & \cellcolor{mygray} \textbf{D0} & \cellcolor{mygray} \textbf{D1} & \cellcolor{mygray} \textbf{D2} \\
 \hline
$\tau=0.5$ & 70.2 & 72.8 & 75.5 \\ 
$\tau=0.3$ & 70.4 & 73.0 & 75.6 \\
$\tau=0.1$ & 70.4 & 72.8 & 75.6 \\ 

\hline
\end{tabular}

\end{table}

\clearpage
%
%

\end{document}